\documentclass[letterpaper]{article} % DO NOT CHANGE THIS
\usepackage{aaai2026}  % DO NOT CHANGE THIS
\usepackage{times}  % DO NOT CHANGE THIS
\usepackage{helvet}  % DO NOT CHANGE THIS
\usepackage{courier}  % DO NOT CHANGE THIS
\usepackage[hyphens]{url}  % DO NOT CHANGE THIS
\usepackage{graphicx} % DO NOT CHANGE THIS
\usepackage{natbib}  % DO NOT CHANGE THIS AND DO NOT ADD ANY OPTIONS TO IT
\usepackage{caption} % DO NOT CHANGE THIS AND DO NOT ADD ANY OPTIONS TO IT
\usepackage{algorithm}
\usepackage{algorithmic}

\usepackage{xcolor}
\usepackage{amsmath}
\usepackage{amsfonts}
\usepackage{amsthm}
\usepackage{tikz}

\newtheorem{theorem}{Theorem}
\newtheorem{definition}{Definition}

\newtheorem{proposition}{Proposition}

\usepackage{makecell}
\usepackage{booktabs}
\usepackage{multicol}
\usepackage{multirow}

\usepackage{newfloat}
\usepackage{listings}
\DeclareCaptionStyle{ruled}{labelfont=normalfont,labelsep=colon,strut=off} % DO NOT CHANGE THIS
\floatstyle{ruled}
\newfloat{listing}{tb}{lst}{}
\floatname{listing}{Listing}
\title{Learning Subgroups with Maximum Treatment Effects without Causal Heuristics}
\author{
Lincen Yang\textsuperscript{\rm1}, Zhong Li\thanks{corresponding author}\textsuperscript{\rm1, \rm2}, Matthijs van Leeuwen\textsuperscript{\rm1}, Saber Salehkaleybar\textsuperscript{\rm1}
}
\affiliations{
    \textsuperscript{\rm1} Leiden Institute of Advanced Computer Science (LIACS), Leiden University \\
    \textsuperscript{\rm2} The Intelligent Computing Research Center, Great Bay University
    \{l.yang, z.li, m.van.leeuwen, s.salehkaleybar\}@liacs.leidenuniv.nl
}

\begin{document}

\maketitle

\begin{abstract}
Discovering subgroups with the maximum average treatment effect is crucial for targeted decision making in domains such as precision medicine, public policy, and education. While most prior work is formulated in the potential‑outcome framework, the corresponding structural causal model (SCM) for this task has been largely overlooked. In practice, two approaches dominate. The first estimates pointwise conditional treatment effects and then fits a tree on those estimates, effectively turning subgroup estimation into the harder problem of accurate pointwise estimation. The second constructs decision trees or rule sets with ad‑hoc `causal' heuristics, typically without rigorous justification for why a given heuristic may be used or whether such heuristics are necessary at all.

We address these issues by studying the problem directly under the SCM framework. Under the assumption of a partition-based model, we show that optimal subgroup discovery reduces to recovering the data-generating models and hence a standard supervised learning problem (regression or classification). This allows us to adopt \emph{any} partition-based methods to learn the subgroup from data. 
% hence, to promote the general approach of reformulating the subgroup discovery task in this context as a regression or classification task, 
We instantiate the approach with CART, arguably one of the most widely used tree-based methods, to learn the subgroup with maximum treatment effect. Finally, on a large collection of synthetic and semi‑synthetic datasets, we compare our method against a wide range of baselines and find that our approach, which avoids such causal heuristics, more accurately identifies subgroups with maximum treatment effect. Our source code is available at \url{https://github.com/ylincen/causal-subgroup}. 
\end{abstract}

% typically formulated in the potential‑outcome framework, rely on ad‑hoc causal heuristics—for example, first estimating individual treatment effects and then running a tree on those estimates—

\section{Introduction}
% Subgroup discovery is the task of learning a subgroup---typically described by simple rules---from data such that the distribution of a target quantity deviates from the whole dataset~\citep{atzmueller2015subgroup}. 
% In the context of causal inference, the quantity of interest is a treatment effect, and it is often desirable to learn a subgroup whose treatment effect is enhanced relative to the overall population or some baseline \citep{su2009subgroup, wang2022causal}. Such analyses have proved useful in applications including healthcare \citep{lipkovich2017tutorial, rothwell2005subgroup, loh2019subgroup} and education \citep{athey2019estimating}.
Subgroup discovery is the task of learning a subgroup, typically described by interpretable rules, from data such that the distribution of a target quantity deviates from that of the full dataset \citep{atzmueller2015subgroup}. In causal inference, the quantity of interest is often the treatment effect, and many works aim to learn subgroups whose effects are enhanced relative to the overall population or a baseline \citep{su2009subgroup, wang2022causal}. Such analyses have proved useful in application domains such as healthcare \citep{lipkovich2017tutorial, rothwell2005subgroup, loh2019subgroup} and education \citep{athey2019estimating}.

In this paper we focus on discovering the subgroup with the maximum average treatment effect, which we name as the \emph{maximum-effect subgroup}, not merely a subgroup with an enhanced or above-average effect. Rather than to characterize the heterogeneity of the treatment effects for the whole population in general, we aim to ask for a single optimum, which can answer questions like which subgroup of patients can benefit most from a certain disease treatment plan~\citep{zhang2017mining, goligher2023heterogeneous, nagpal2020interpretable}. 
Importantly, this maximum-effect objective also provides a foundation for understanding heterogeneity more generally: one can iteratively remove the discovered subgroup and re-apply the method to the remaining instances—an approach often referred to as sequential covering or divide-and-conquer \citep{furnkranz2012foundations, cohen1995ripper}.

Existing methods for learning subgroups with enhanced treatment effects from data largely take the following two approaches. 
The first approach takes a two-step process~\citep{foster2011subgroup, huang2025distilling}: 1) estimating the pointwise conditional treatment effect, and 2) applying an off-the-shelf subgroup discovery method~\citep{lavravc2004subgroup, van2012diverse} or fitting a partition-based model, e.g., a classification tree~\citep{breiman1984classification} or a rule set~\citep{clark1989cn2}, to the estimated pointwise effects to obtain the subgroups with enhanced treatment effects. While flexible, this strategy converts subgroup discovery into the arguably harder task of accurately estimating pointwise conditional treatment effects, and the learned subgroups can be highly sensitive to the pointwise estimation error.

The second approach directly tries to search for subgroups with enhanced treatment effects~\citep{zhou2024curls, dusseldorp2014qualitative, athey2016recursive, su2009subgroup}. It estimates the subgroup treatment effect (i.e., the treatment effect conditioned on the subgroup) and uses that estimate to design the heuristics for searching the subgroups. Algorithmically, the search can be achieved by either a tree-based approach, i.e., grow a tree and pick a subgroup from the leaves~\citep{dusseldorp2014qualitative, su2009subgroup}, or a rule-based approach that typically reduces to a combinatorial optimization problem~\citep{zhou2024curls}. In practice, these methods often blend the estimated subgroup treatment effect with regularization terms (based on the number of instances contained in the subgroups). 
However, the proposed heuristics typically lacked theoretical justification at the time they were introduced, and, when the goal is to find the maximum-effect subgroup, two issues arise:
% While this might be reasonable for enhanced-effect discovery, it raises two questions when the target is the maximum: 
1) while a large number of different heuristics have been proposed, which heuristic, if any, is theoretically appropriate for identifying the maximizer, and 2) are specialized `causal' heuristics necessary at all? 

Thus, these questions hinder the naive approach of using existing methods that discover the subgroups with enhanced treatment effect and then pick the single subgroup within them as the maximum-effect subgroup. 

To tackle these shortcomings, we leverage the structural causal model (SCM) framework, and prove that, under the assumption of a partition-based model, the task of learning the maximum-effect subgroup can be reduced to a standard supervised learning task that aims to reveal the data-generating model. We instantiate this approach by training a CART~\citep{breiman1984classification} tree, one of the most commonly used methods for supervised learning. We empirically compare against commonly used baseline methods, using both synthetic and semi-synthetic datasets. We demonstrate that our approach, even with the classic yet simple CART algorithm, shows superior performance in the task of maximum-effect subgroup discovery. To our knowledge, this is the first work to ground maximum-effect subgroup discovery within structural causal inference via rigorous theoretical results.

It is noteworthy to mention that our primary contribution is \emph{not} to propose a new, specific heuristic and/or algorithm for discovering subgroups with maximum treatment effects; instead, we provide a theoretical result that offers insights into whether `causal' heuristics are unnecessary for the task of subgroup discovery in this context. Our empirical results are aimed at justifying the general approach of reducing the task of discovering the maximum-effect subgroup to the standard supervised machine learning tasks.

\section{Related Work}
 % tree-based
Existing methods for discovering subgroups with maximum/enhanced treatment effects can be categorized into three approaches. 

The first approach aims to learn a decision tree or a decision rule set, with specifically designed learning criteria and/or heuristics that are tailored for causal inference. 
% The criteria/heuristics often consist of the empirical subgroup treatment effect, which is estimated by first extracting the subgroup of instances satisfying the condition of the corresponding rule, and then using the empirical average (or the empirical proportions of instances with the target value equal to 1) as the estimate. 
These criteria/heuristics often combine an empirical estimate of the subgroup treatment effect---computed by selecting instances that satisfy the rule that describes this subgroup and averaging outcomes within treatment and control---with regularization terms. 
% For instance, the seminal work of Causal Tree (CT)~\citep{athey2016recursive} proposed to empirically calculate the estimated average treatment effect for the subset of instances contained in any tree node, and use the mean-squared error (MSE) of the estimated treatment effect as the splitting and pruning criteria (since the ground-truth of the conditional treatment effect for each individual point is not observed due to the lack of the counterfactual outcome, an unbiased estimator of the MSE is proposed as an approximation). 
For instance, the seminal work Causal Tree~\citep{athey2016recursive} estimates the average treatment effect within each node and uses an unbiased proxy for the mean-squared error of the treatment‐effect estimator for splitting and pruning, since individual counterfactual outcomes are unobserved.
Further, QUINT~\citep{dusseldorp2014qualitative} adopts the weighted sum of the subgroup treatment effect of each leaf node plus a term that encourages large subgroup size as the splitting criterion. As QUINT uses a bootstrap procedure for pruning with an expensive computational cost, the scalability is limited. Similarly, Interaction Tree~\citep{su2009subgroup} uses the difference between the subgroup treatment effects of two children nodes (standardized by the estimated pooled variance) as the splitting criterion. 

% rule-based
Beyond trees, SIDES~\citep{lipkovich2011subgroup} directly searches for subgroups without explicitly building a tree. SIDES provides several heuristics to split a subgroup into more refined subgroups, and essentially leaves the choice of which heuristic to use to the users. Last, the more recent rule set methods, including CURLS~\citep{zhou2024curls} and MOSIC~\citep{chen2025mosic}, adopt a similar criterion in learning a rule set and directly maximize the subgroup treatment effect, regularized by a model complexity term.

Thus, we conclude that although various criteria have been proposed, there is still no principled guideline for determining which is most appropriate to use in a given practical scenario. Our theoretical analysis and superior empirical experiment results both challenge the necessity of leveraging such learning objectives and/or algorithmic heuristics to learn the maximum-effect subgroup. 

The second approach first estimates pointwise conditional treatment effects and then partitions the feature space using those estimates. Specifically, Virtual Twins~\citep{foster2011subgroup} first builds a predictive model to predict the counterfactual outcomes by setting the treatment variable to zero or one. Then, it fits a tree or regressor to produce approximately homogeneous-effect regions. 
% Then, it uses the predicted outcomes to obtain an estimated conditional treatment effect for each single data point. Last, a decision or regression tree is fit to partition the feature space into subsets with (approximately) homogeneous treatment effect. 
Similarly, the X-learner~\citep{kunzel2019metalearners} first splits the dataset into the treatment (i.e., the treatment variable $T=1$) and the control group ($T=0$), then separately learns two models to predict the conditional treatment effect of individual points, and finally applies any supervised method to model the pointwise conditional treatment effects. Further, Distill Tree~\citep{huang2025distilling} can leverage any off-the-shelf method that predicts pointwise conditional treatment effect, and then use a `student model' to distill the subgroups. The first step here is often referred to as the task of (individual) conditional treatment effect estimation, which can be achieved by various approaches, including double machine learning~\citep{chernozhukov2018double}, meta-learners~\citep{nie2021quasi, kunzel2019metalearners}, (ensemble) tree methods~\citep{athey2019generalized, hahn2020bayesian}, and (deep) representation learning~\citep{curth2021inductive, shi2019adapting, lee2025subgroupte} (although not all these methods have been considered in learning subgroups with the maximum or enhanced treatment effects). 

We argue that estimating pointwise conditional treatment effect is a much harder problem than estimating the subgroup treatment effect in practice; hence, the variance of the estimator is in general large and will highly depend on the chosen model. 

The third approach is model-based recursive partitioning~\citep{seibold2016model}. Unlike the previous two families, it does not treat the subgroup effect as constant. Instead, it starts with a global predictive model that associates the target variable $Y$ with the feature variables $X$ and treatment variable $T$. It then iteratively splits the feature space into hyper-cubes, and refits model parameters within each region. However, as shown by our theoretical results, a subgroup that maximizes the average treatment effect must exhibit homogeneous pointwise treatment effects (which we explain in detail later). By design, model-based approaches allow within-subgroup variation and therefore do not target the maximum-effect subgroup as defined here.

% a subgroup that maximizes the treatment effect must have homogeneous pointwise conditional treatment effects (which we explain in detail later), a model-based approach does not fulfill our goal of subgroup discovery. This is because a model-based approach, although more flexible and expressive, by definition does not have homogeneous pointwise estimates. 

% \begin{itemize}
%     % \item Hypothesis testing
%     % \item recursive partitioning
%     % \item from individual CATE to subgroup ATE. 
%         % \item use a summary table? with columns like: need Individual CATE, hypothesis testing, causal heuristics for splitting/growing.. 
%     \item TODO: 
%     \item model-based (We would like to avoid heterogeneous treatment effect within the subgroup)
%     \item Subgroup discovery in the non-causal setting?
%     \item (make this more concise and it might be useful in the Introduction: Existing methods that search for subgroups with maximized/enhanced treatment effects often adopt the potential outcome framework~\citep{??}, together with the standard assumptions of observational inference including the conditional ignorability~\citep{??} which is equivalent to the no-hidden-confounder assumption in the SCM framework. )
% \end{itemize}

\section{Theory}
We first review the basic concept of structural causal models and the definition of `rules'. Then, we show that the maximum-effect subgroup must have a homogeneous pointwise conditional treatment effect. Last, we introduce the partition-based model and present our main theorem, which states that learning the maximum-effect subgroup reduces to the task of learning the underlying data-generating process. 
\subsection{Preliminaries}
% \textcolor{red}{To do/To decide: Depending on the space left, we might want to include a brief discussion on basic concepts like SCM, identifiability, the "do" notation, graphic model, etc?}
\noindent
\textbf{Structural causal models.}
A structural causal model (SCM) is a set of functions that fully specifies the data-generating process: given a set of variables $\{V_i\}_{i \in \{1, .., m\}}$, a SCM can be specified by a set of functions $V_i = f_i(pa(V_i), U_i)$, in which $pa(V_i)$ denotes the parent nodes of $V_i$ and all $U_i$ are independent~\citep{pearl2009causality}. The causal relationship is often represented as a directed acyclic graph (DAG) with nodes $\{V_i\}$ where there is a directed edge from $V_j$ to $V_i$ if $V_j\in pa(V_i)$. Given an SCM, we can specify an intervention distribution, denoted as $P(V_i|do(V_j=v_j))$, as the conditional probability of $V_i$ specified by the SCM model after replacing the original equation $V_j = f_i(pa(V_j), U_j)$ by $V_j = v_j$. 

Under certain conditions \cite{pearl2012calculus}, the intervention distribution $P(V_i|do(V_j=v_j))$ can be computed from expressions written based on the observational distribution. This is often referred to as `identifiability', and can be used to estimate $P(V_i|do(V_j=v_j))$ from the observational data. 

% To estimate the intervention distribution from observational data, one can resort to the so-called identifiability, 
\smallskip \noindent 
\textbf{Subgroups and rules.}
A subgroup is often described by a rule, which is a logical conjunction of literals~\citep{van2012diverse, zhou2024curls}: given a dataset with feature variables $X_j, j \in \{1, ..., m\}$, a literal is in the form of $X_j \in R_j$, in which $X_j$ is a single feature variable and $R_j$ represents a subset of the domain of $X_j$; hence, a rule can be denoted as a conjunction of several literals $\bigwedge X_j \in R_j$. 
\subsection{Causal subgroup discovery}
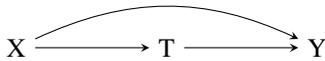
\begin{figure}[ht]
\centering
\begin{tikzpicture}[
  ->,
  >=stealth,
  node distance=2cm
]

\node (X) {X};
\node (T) [right of=X] {T};
\node (Y) [right of=T] {Y};

\draw (X) -- (T);
\draw (T) -- (Y);
\draw[->] (X) to[bend left=25] (Y);

\end{tikzpicture}
\caption{The DAG for the SCM for treatment effect estimations.}\label{fig:dag1}
\vspace{-3mm}
\end{figure}

We first consider the general form of structural causal model (SCM) for estimating treatment effect, i.e., 
\begin{align*}
    X = N_X, T = f_T(X, N_T), Y = f_Y(T, X, N_Y), \\
    N_X, N_T, N_Y \text{  are independent exogenous noises,}
\end{align*}
\noindent with the DAG as shown in Fig.~\ref{fig:dag1}.

$X$ can be a multivariate random variable with its domain denoted as $\mathcal{X}$. Further, we also assume that there are no hidden confounders that can affect both the treatment variable $T$ and the target variable $Y$. \footnote{Previous methods for the task of subgroup discovery for enhanced/maximum treatment effects~\citep{athey2016recursive, dusseldorp2014qualitative, su2009subgroup, wang2022causal, zhou2024curls} often leverage the potential outcome framework~\citep{Rubin1974}, and they often have the “exchangeability”~\citep{RosenbaumRubin1983} assumption, which is equivalent to the no hidden‑confounder assumption.}
% \footnote{Previous methods for the task of subgroup discovery for enhanced/maximum treatment effects often leverage the potential outcome framework~\citep{??}, and they often have the ``exchangeability"~\citep{??} assumption, which is equivalent to our no-hidden-confounder assumption.} 

To simplify notations, we assume that $X$ is discrete and $Y$ is a binary target variable \emph{when developing our theory}. However, our theoretical results can be directly extended to the case when $X$ is continuous, by replacing the $\sum$ with the integral sign. The results can also be extended to numeric targets, by simply replacing the $P(.)$ with $E(.)$ in our derivations. Our experiments do contain datasets with continuous $X$ and $Y$.

Our goal is to find a subgroup such that the treatment effect of the subgroup is maximized. Specifically, we define the \emph{subgroup treatment effect} as follows.
\begin{definition}[Subgroup treatment effect] Given a $Q \subseteq \mathcal{X}$, in which $\mathcal{X}$ is the domain of $X$, we say $Q$ is a subgroup, and meanwhile we define the subgroup treatment effect as
\begin{align} \label{eq:te_def}
A(Q) := P(Y=1|do(T:=1), X \in Q) - \notag \\ P(Y=1|do(T:=0), X \in Q).
\end{align}
\end{definition}
\noindent 
We hence formally state our problem as finding the maximum-effect subgroup 
\begin{equation}
    Q^* = \arg \max_Q A(Q).
\end{equation}
Further, under the assumption that there exists no hidden confounder and $P(X\in Q) > 0$ (i.e., subgroups with probability zero are not interesting in practice), $A(Q)$ is \emph{identifiable} from the observational distribution: 
% Further, we can derive $A(Q)$ as the expectation of conditional treatment effect defined as $A(X):= P(Y=1|do(T:=1), X) - P(Y=1|do(T:=0), X)$, as shown in the proposition below (we defer the proof to the supplementary materials).
\begin{proposition}[Identifiable] \label{prop1}
$A(Q) = E_{X|X \in Q} [P(Y=1|T=1, X) - P(Y=1|T=0, X)]$ under the assumptions stated above.
\end{proposition}
\noindent The proof is deferred to the supplementary materials. 
% \begin{proof}
% {\small
%     \begin{align}
%         &P(Y=1 |do(T:=1), X \in Q) - \notag \\
%         & P(Y=1 |do(T:=0), X \in Q) \notag \\
%         = & \sum_{x \in Q}P(Y=1|do(T:=1), X=x) P(X=x|X\in Q) - \notag \\ 
%          &  \sum_{x \in Q}P(Y=1|do(T:=0), X=x) P(X=x|X\in Q) \label{eq:use_insertion_rule}\\
%         = & E_{X|X \in Q} A(X) \notag
%     \end{align}
% }
% \end{proof}

Next, we can further prove that an maximum-effect subgroup $Q^*$ must have homogeneous treatment effect, which we formally define as follows:
\begin{definition} [Subgroup with a homogeneous treatment effect] \label{def:homo_treatment}
    A subgroup $Q \subset \mathcal{X}$ has homogeneous treatment effect if $\forall Q' \subset Q$, $A(Q) = A(Q')$. Equivalently, as $Q'$ can be a subset that contains only one single point, i.e., $Q' =\{x'\}$ for some $x' \in \mathcal{X}$, we also have $A(Q) = A(X=x), \forall x \in Q$, in which $A(X=x) := P(Y|do(T:=1), X=x) - P(Y|do(T:=0), X=x)$.
\end{definition}
The next theorem justifies the approach of finding the maximum-effect subgroup through subgroups with homogeneous treatment effects. 
\begin{theorem} \label{thm1}
For a subset $Q' \subset Q$, if $A(Q') \leq A(Q)$ then we must have $A(Q \setminus Q') \geq A(Q)$, and the equality $A(Q \setminus Q') = A(Q)$ only holds when $A(Q') = A(Q)$.
\end{theorem}
\begin{proof}
    \begin{align} \label{eq:thm1}
        A(Q) & \stackrel{}{=} \sum_{x\in Q} (P(Y=1|do(T:=1), X=x) - \notag \\
        & P(Y=1|do(T:=0), X=x)) P(X=x|X\in Q) \notag\\
        = & \sum_{x \in Q' }(P(Y=1|do(T:=1), X=x) - \notag\\
        & P(Y=1|do(T:=0), X=x)) P(X=x|X\in Q ) + \notag \\
        & \sum_{x \in Q\setminus Q' }(P(Y=1|do(T:=1), X=x) - \notag \\
        & P(Y=1|do(T:=0), X=x)) P(X=x|X\in Q )
    \end{align}
    Note that $\forall x \in Q'$, we have $P(X=x) = P(X=x, X\in Q')$, and as a result
    \begin{align*}
        & P(X=x|X\in Q ) = \frac{P(X=x, X\in Q)}{P(X \in Q)} = \frac{P(X=x)}{P(X \in Q)} \\
        = & \frac{P(X=x)P(X\in Q')}{P(X\in Q) P(X\in Q')} 
        = P(X=x|X\in Q') \frac{P(X\in Q')}{P(X \in Q)};
    \end{align*}
    Similarly, we can also show that
    \begin{align*}
        P(X=x|X\in Q ) = P(X=x|X\in Q\setminus Q') \frac{P(X\in Q \setminus Q')}{P(X \in Q)}.
    \end{align*}
    Thus, by substituting $P(X=x|X\in Q )$ in Eq.~\ref{eq:thm1}, we have 
    \begin{align*}
        A(Q) & = \frac{P(X\in Q')}{P(X \in Q)} A(Q') + \frac{P(X\in Q\setminus Q')}{P(X \in Q)} A(Q\setminus Q').
    \end{align*}
    Hence
    \begin{equation}
        \min\{A(Q', Q\setminus Q'\} \leq A(Q) \leq \max\{A(Q', Q\setminus Q'\};
    \end{equation}
    and the equality only holds when $A(Q') = A(Q\setminus Q')$.
\end{proof}
Thus, an maximum-effect subgroup $Q^*$ must have homogeneous treatment effect (otherwise the subgroup can be further split and one of the subsets will have a higher average treatment effect).

\subsection{Our proposed model}
% A sufficient condition for achieving homogeneous treatment $A(Q)$ is by leveraging a partition-based model, which we formally define by the following structural causal model (SCM): 
We propose a partition-based model, which we formally define by the following structural causal model (SCM): 
% that is, we assume a partition of the domain of $X$, denoted as $\{K_i\}_{i \in I}$, in which  $K_i \subseteq \mathcal{X} = \mathbb{R}^m, \forall i, j \in I, K_i \cap K_j = \emptyset$, such that the functional relationship between $X$ and $Y$ fully depends on the indicator function defined based on the partition. 
% However, with finite sample size in practice, we also need to regularize the model in order to have reliable estimates for the conditional probability $P(Y|X \in Q, T=t), t\in\{0, 1\}$ for a given subgroup $\{X \in Q\}$ for which we would like to estimate the treatment effect. 
% \begin{align*}
%     X = N_X, T = f_T(X, N_T), Y = f_Y(T, S, N_Y) \\
%     S = \sum_i^I i \cdot \mathbf{1}_{K_i}(X), K_i \subseteq \mathbb{R}^m, \forall i, j \in [I], K_i \cap K_j = \emptyset, \\
%     N_X, N_T, N_Y \text{  are independent exogenous noises.}
% \end{align*}
\begin{definition} \label{def:model}
The partition-based model is defined as 
    \begin{align*}
    X = N_X, T = f_T(X, N_T), \\ 
    Y = f_Y(T, \sum_i^I i \cdot \mathbf{1}_{K_i}(X), N_Y)\\
    K_i \subseteq \mathcal{X}, \forall i, j \in I, K_i \cap K_j = \emptyset, \\
    N_X, N_T, N_Y \text{  are independent exogenous noises,} 
\end{align*}
in which $\mathbf{1}_{K_i}(.)$ is the indicator function which is equal to one if $X\in K_i$ (otherwise, it is zero), and $I = \{1, 2, ..., |I|\}$ is an index set. The causal relationship among variables can be represented by Figure~\ref{fig:dag1} as well. 
\end{definition}
In the partition-based model, we assume the feature space can be partitioned into small subsets, within each the conditional probability $P(Y|X, do(T=t)), t\in\{0, 1\}$ becomes homogeneous; i.e., $\forall x \in K_i, P(Y|X=x, do(T=t)) = P(Y|X \in K_i, do(T=t))$. 

Although most subgroup discovery methods adopt the potential outcome framework~\citep{Rubin1974}, we argue that they often implicitly assume this partition-based model. Specifically, when predicting the subgroup treatment effects, all instances in the same subgroup (or a tree leaf node for tree-based methods) are assigned with a single predicted outcome for $T=t$, where $t\in \{0,1\}$~\citep{athey2016recursive}. 

% We acknowledge that this is not a necessary condition as well; however, without assuming the partition-based model and homogeneous conditional probability, the common approach would be to estimate the subgroup treatment effect $A(Q)$ by first estimating $A(X=x)$ (defined in Def.~\ref{def:homo_treatment}) for all possible values $x \in \mathcal{X}$, and then take the average by leveraging Proposition~\ref{prop1}~\citep{huang2025distilling, foster2011subgroup}. However, estimating point-wise conditional treatment effect $A(X=x)$ is a much harder problem. We also compared against baseline methods that adopt the approach of averaging point-wise conditional treatment effect $A(X=x)$ in the Experiment section later, and demonstrate our superior performance. 

% Note that this is not only the common assumptions in classification/regression trees, but also the implicit assumptions when assuming $P(Y|X, T=t), t\in\{0, 1\}$ can be estimated from the empirical proportions of $Y=y$ within a given subgroup. 

We next show that under the condition that the data is generated according to the partition-based model (Def.~\ref{def:model}), the subgroup that maximizes the treatment effect, defined in Eq.~\ref{eq:te_def}, is one of the $K_i, i\in I$. 

\begin{theorem} \label{thm:main}
Denote $S = \sum_{i\in I} i \cdot \mathbf{1}_{K_i}(X), K_i \subseteq \mathcal{X}, \forall i, j \in I, K_i \cap K_j = \emptyset$. Without loss of generality, assume that
{\small
    \begin{align} \label{eq:assumption2}
       & P(Y=1|S=1, do(T:=1)) - P(Y=1|S=1, do(T:=0)) \geq \notag \\ 
       & P(Y=1|S = i, do(T:=1)) - P(Y=1|S=i, do(T:=0)) \notag \\ 
       & (\forall i \in \{2, ..., n\}) . 
    \end{align} 
}
Then, $\forall Q \subseteq \mathcal{X}$, we have
{\small
    \begin{align} \label{eq:thm2}
        P(Y=1|S=1, do(T:=1)) - 
        P(Y=1|S=1, do(T:=0)) \geq \notag \\ P(Y=1|X\in Q, do(T:=1)) -  P(Y=1|X\in Q, do(T:=0))
    \end{align}  
}
\end{theorem}
\begin{proof}
Without loss of generality, we assume that $S \neq 0$ no matter what value $X$ takes. This can be achieved by making the last subgroup $K_n = \mathcal{X} \setminus (\cup_{i \in \{1, ..., n-1\}} K_i)$. 
The right hand side (RHS) of Eq.~\ref{eq:thm2} is 
\begin{align*}
 \text{RHS} & \stackrel{\text{(a)}}{=} \frac{ P(Y=1, X \in Q| do(T:=1))}{ P(X \in Q| do(T:=1))} -  \,\,\,\,\,\,\,\,\,\,\,\,  \,\,\,\,\,\,\,\,\,\,\,\,\,\,\,\,\,\,\,\,\,\,\,\,   \,\,\,\,\,\,\,\,\,\,\,\,\,\,\,\,\,\,\,\,\,\,\,\,\\
    & \frac{ P(Y=1, X \in Q| do(T:=0))}{ P(X \in Q| do(T:=0))} \\
      & \stackrel{\text{(b)}}{=} \frac{ \sum_{i} P(Y=1, X \in Q\cap K_i| do(T:=1))}{ P(X \in Q| do(T:=1))} - \\
      & \frac{  \sum_{i} P(Y=1, X \in Q\cap K_i| do(T:=0))}{ P(X \in Q| do(T:=0))} \\
      &  = \sum_{i} \Bigl(\frac{P(X \in Q \cap K_i | do(T:=1))}{P(X\in Q |do(T:=1))} \cdot \\
      & P(Y=1|X\in Q\cap K_i, do(T:=1)) \Bigr) - \\
      & \sum_{i}  \Bigl( \frac{P(X \in Q \cap K_i | do(T:=0))}{P(X\in Q |do(T:=0))} \cdot \\
      & P(Y=1|X\in Q\cap K_i, do(T:=0))  \Bigr) \\ 
      & \stackrel{\text{(c)}}{=} \sum_i \Bigl(P(Y=1|X \in Q \cap K_i, do(T:=1)) - \notag\\
      & P(Y=1|X \in Q \cap K_i, do(T:=0)) \Bigr) \frac{P(X \in Q \cap K_i)}{P(X \in Q)} \notag \\
      \end{align*}
    \begin{align*}
       & \stackrel{\text{(d)}}{=}   \sum_i \frac{P(X \in Q \cap K_i)}{P(X \in Q)} \bigl( P(Y=1|S=i, do(T:=1)) -\notag \\
       & P(Y=1|S=i, do(T:=0))\bigr)  \notag\\
       & \stackrel{\text{(e)}}{\leq} \sum_i \frac{P(X \in Q \cap K_i)}{P(X \in Q)} \bigl( P(Y=1|S=1, do(T:=1)) - \notag\\ 
       & P(Y=1|S=1, do(T:=0))\bigr)  \notag \\
       & = \text{LHS of Eq.~\ref{eq:thm2}}, \notag 
\end{align*}
where in:
\\
(a) we use the Bayes rule;\\ 
(b) we partition the event $\{X \in Q\}$ into subsets $\{X \in Q \cap K_i\}_{i \in I}$; \\
(c) we use the rule of `deletion of intervention' of do-calculus~\citep{pearl2009causality}; \\
(d) we leverage our partition-based model property that since $Y = f_Y(T, \sum_i^I i \cdot \mathbf{1}_{K_i}(X), N_Y)$, $P(Y|S=i, do(T=t)) = P(Y|X\in K_i, do(T=t)) = P(Y|X\in K_i\cap Q, do(T=t))$; \\
(e) we use the assumption Eq.~\ref{eq:assumption2}.
% Note that Eq.~\ref{eq:proof_delete_do_calc} is obtained by the `deletion' rule of do-calculus~\citep{pearl2009causality}.

\noindent (\textbf{Comments:} We also extend Theorem~\ref{thm:main} to the case with hidden confounders in the supplementary materials. )
\end{proof}

% Theorem~\ref{thm:main} suggests that searching for the subgroup that maximize the treatment effect is equivalent to the following two steps: 1) revealing the data-generating model, which can be achieved by simply maximizing the likelihood, or equivalently, optimizing the cross-entropy loss (or approximately achieving the goal via a surrogate loss like Gini-index), and 2) examining all the subgroups and search for the one that maximize the treatment effect. 
% That is, Theorem~\ref{thm:main} states that, given a partition-based model that partitions the feature space of $X$ into a collection of subsets $\{K_i\}_{i \in I}$, the subgroup $Q^*$ that maximizes the subgroup treatment effect $A(Q^*)$ must be one of the $K_i, i\in I$, i.e., $Q^* = \arg\max_{i \in I} A(K_i)$. 
% Therefore, Theorem~\ref{thm:main} suggests that searching for the subgroup that maximize the treatment effect is equivalent to the following two steps: 1) revealing the data-generating model, which can be achieved by simply maximizing the likelihood, or equivalently, optimizing the cross-entropy loss (or approximately achieving the goal via a surrogate loss like the Gini-index as used in the CART decision tree algorithm~\citep{breiman1984classification}), and 2) examining all the subgroups and search for the one that maximize the treatment effect. 
% In practice, the partition-based model learned from data typically contains significantly fewer subgroups than the sample size to avoid overfitting, making it efficient to iterate through all the learned subgroups.

Theorem~\ref{thm:main} states that, under a partition-based model that divides the feature space into subsets $\{K_i\}_{i\in I}$, the subgroup $Q^*$ that maximizes the subgroup treatment effect must be one of the partition subset, i.e., $Q^* = \arg\max_{i \in I} A(K_i)$.
% \[
% Q^* \in \{K_i\}_{i\in I}
% \qquad\text{and}\qquad
% Q^* \;=\; \arg\max_{i\in I} A(K_i).
% \]
Consequently, searching for the maximum-effect subgroup reduces to two steps: 1) recovering the data-generating partition, e.g., by maximum likelihood (equivalently, minimizing cross-entropy) or a surrogate such as the Gini index used by CART~\citep{breiman1984classification}, and 2) evaluating the learned subsets and selecting the one with the largest estimated treatment effect. In practice, the learned partition typically contains far fewer subsets than the sample size (to avoid overfitting), so scanning all subsets is computationally lightweight.

In sum, our theorem questions the necessity of using bespoke `causal' heuristics and/or learning criteria for discovering maximum-effect subgroups. Instead, Theorem~\ref{thm:main} indicates that we can use any off-the-shelf decision-tree-based or rule-based methods for classification (for nominal $Y$) or regression (for continuous $Y$) for such task. 

\section{Learning the Subgroup from Data}
As our Theorem~\ref{thm:main} does not restrict us to use any specific learning method, we instantiate our approach with CART \citep{breiman1984classification}, a widely used tree method for classification and regression. Notably, rather than proposing a specific, new algorithm for learning maximum-effect subgroups, we aim to empirically validate the general approach of learning such subgroups by reducing it to a standard classification or regression task. Hence, other advanced tree/rule-based methods can be adopted here as well~\citep{yang2022truly,yang2024conditional,hu2019optimal,brita2025optimal}.

% we empirically use CART~\citep{breiman1984classification} as it is the most popular tree-based method in supervised machine learning tasks including classification and regression. 
% Crucially, we aim to challenge the necessity of designing causal-specific subgroup discovery method---often with empirical estimates of the subgroup treatment effects as the learning criterion and/or algorithmic heuristics (e.g., the splitting criterion for tree-based methods). 
% Thus, instead of designing a new supervised tree/rule-based method, we leverage the arguably most widely used off-the-shelf method to validate our proposed approach. 
% tree-based method for obtaining the subgroup with the maximum treatment effect. 

% Notably, we do include the treatment variable $T$ in the training phase: as we assume the target is a function of $T$ and the sum of an indicator function of feature vector $X$, for the subgroup with the maximum treatment effect, the conditional distribution of $Y$

Specifically, for a discrete target variable $Y$, we fit a CART decision tree, together with the standard criterion Gini-index. Gini-index is a surrogate loss of cross-entropy loss, and hence an approximation for maximizing the likelihood. Meanwhile, for a continuous target variable, we fit a CART regression tree, with the standard mean squared error (MSE) as the learning criterion~\citep{breiman1984classification}. For numeric targets, optimizing MSE corresponds to maximum likelihood under the assumption that the conditional distribution of the target variable is Gaussian. 
% For a numeric target $Y$, optimizing MSE for revealing the data-generating model is equivalent to assuming the conditional distribution of the target variable to be Gaussian. 
To mitigate overfitting, we apply the common approach of cost-complexity tree pruning with the help of cross-validation.

After the decision (regression) tree is learned from data, we next describe how to find the subgroup with the maximum treatment effect. Given any leaf node of the learn tree, the path from the root node to this leaf node can be considered a single rule $R$. Hence, each rule defines a subgroup (hyper-rectangle) of the feature space $\mathcal{X}$ which we denote as $Q$. Notably, if rule $R$ contains a condition of $T=t, (t\in \{0,1\})$, the subgroup is obtained by ignoring the internal node that contains the condition for the treatment variable $T$. 
% By contrast, if such rule $R$ does not contain a condition of $T=t, (t\in \{0,1\})$, the predictions for $P(Y|T=t, X\in K), t\in \{0,1\}$ would be the same, indicating a zero subgroup treatment effect. Thus, we can ignore such subgroups as we 
% If such rule $R$ contains a condition of $T=t, (t\in \{0,1\})$, it indicates that the prediction of $P(Y=1|T=t, X\in K)$ might be different from that of $P(Y=1|T=1-t, X\in K)$ (as predicting the latter one requires querying the estimations of other one or more leaf nodes), and hence a non-zero estimated treatment effect. That is, each rule given by the root-to-leaf path 

To estimate the subgroup treatment effect of the subgroup $Q$, we adopt `honest' inference that is proposed in the previous work of Causal Tree~\cite{athey2016recursive}. Before we train the decision (regression) tree, we randomly split the dataset into the training and test sets. We first train the decision (regression) tree model with the training set; then, with the model learned from it, we estimate the treatment effect for each subgroup with the test set. Precisely, given the instances of the test set that is contained in the subgroup $Q$, we further split it into two subsets, one with $T=0$ and the other with $T=1$. With these two subsets, we can calculate the empirical conditional probability $\hat{P}(Y|T=t, X\in Q), t\in\{0,1\}$. 

% Next, CART aims to train the decision tree such that each leaf will have approximately homogeneous leaves~\cite{breiman1984classification}; i.e., it aims to learn a tree with leaf nodes $Q$ such that $P(Y|T=t, X\in Q) \approx P(Y|T=t, X=x), t\in \{0,1\}, \forall x \in Q$ (if $Q$ is one of the ground-truth subgroups in the generative model, `$\approx$' can be replaced by `$=$'). Thus, this motivates us to estimate $P(Y|T=t, X=x)$ in Proposition~\ref{prop1} by $\hat{P}(Y|T=t, X\in Q), t\in\{0,1\}$, which leads to our proposed estimator for $A(Q)$ as 
Next, since CART aims to partition the feature space such that each leaf node contains instances with approximately homogeneous conditional probabilities, its estimate of $P(Y|X, T)$ for all instances reaching a particular leaf $Q$ is given by the empirical average of the target values of those instances: e.g., for $T=1$ and a binary target $Y=1$, $\hat{P}(Y=1|X=x,T=1) = \frac{|\{(x,t,y) \in D: y=1, x\in Q, t=1\}|}{|\{(x,t,y) \in D: x\in Q, t=1\}|} =: \hat{P}(Y=1|X\in Q, T=1)$, in which $D$ denotes the dataset and $|.|$ denotes set cardinality. By substituting these empirical estimates into the expression in Proposition~\ref{prop1}, we obtain the proposed estimator for $A(Q)$: 
{\small
\begin{align*}
\hat{A}(Q):= \hat{P}(Y=1|X\in Q, T=1) - \hat{P}(Y=1|X\in Q, T=0).
\end{align*}
}
\noindent Finally, we go over all subgroups and pick the subgroup with the highest estimated treatment effect as our maximum-effect subgroup learned from data, which we denote as $\hat{Q}$. 

\subsection{The advantage over `causal' heuristics} 
Guided by Theorem~\ref{thm1}, one might be tempted to maximize the empirical subgroup treatment effects and use it as a heuristic when learning the tree-based partition model. For instance, when splitting a tree node $Q$ into two children $Q_L$ and $Q_R$, one could regard each tree node as a subgroup (as previously described), and build the heuristics in the form of $\max\{\hat{A}(Q_L), \hat{A}(Q_R)\}$, or $|\hat{A}(Q_L) - \hat{A}(Q_R)|$. These heuristics are commonly used in previous works~\citep{dusseldorp2014qualitative, athey2016recursive, su2009subgroup} as specifically designed `causal' heuristics. 

However, these heuristics are fragile because $\hat A(Q)$ requires accurate estimation of \emph{both} terms in the definition of $A(Q)$ (Eq.~\ref{eq:te_def}). During tree growth, imbalances between the treatment $(T=1)$ and control $(T=0)$ groups are common. For instance, the extreme case would be that there are many data points satisfying the condition $\{T=1 \wedge X \in Q\}$ yet few points satisfying the condition $\{T=0 \wedge X \in Q\}$. 
Under such circumstances, one might prefer to shrink $Q$ to potentially increase $\hat{P}(Y=1|T=1, X \in Q)$, while also being inclined to expand $Q$ such that it contains more instances, in order to have a more reliable estimation for $\hat{P}(Y=1|T=0, X \in Q)$. These objectives are inherently conflicting and cannot be achieved simultaneously. 
% Under such circumstances, one might prefer to shrink $Q$ to potentially increase $\hat A(Q)$ (as suggested by Theorem~\ref{thm1}), while also being inclined to expand $Q$ so that $Q$ would contain more instances, in order to have a more reliable estimation for 

Thus, leveraging $\hat{A}(Q)$ as splitting heuristics either leads to large variance in the training phase and the risk of overfitting, or a compromise between the two subsets of instances with $T=1$ and $T=0$, respectively. 
We avoid this pitfall by optimizing the Gini index, a well-behaved surrogate for cross-entropy (i.e., log-likelihood maximization). This objective remains stable under imbalance and defers the treatment effect estimation to an `honest' stage where both subsets with $(T=1)$ and $(T=0)$ are adequately supported.

\begin{figure}[ht]
    \centering
    \includegraphics[width=0.95\linewidth]{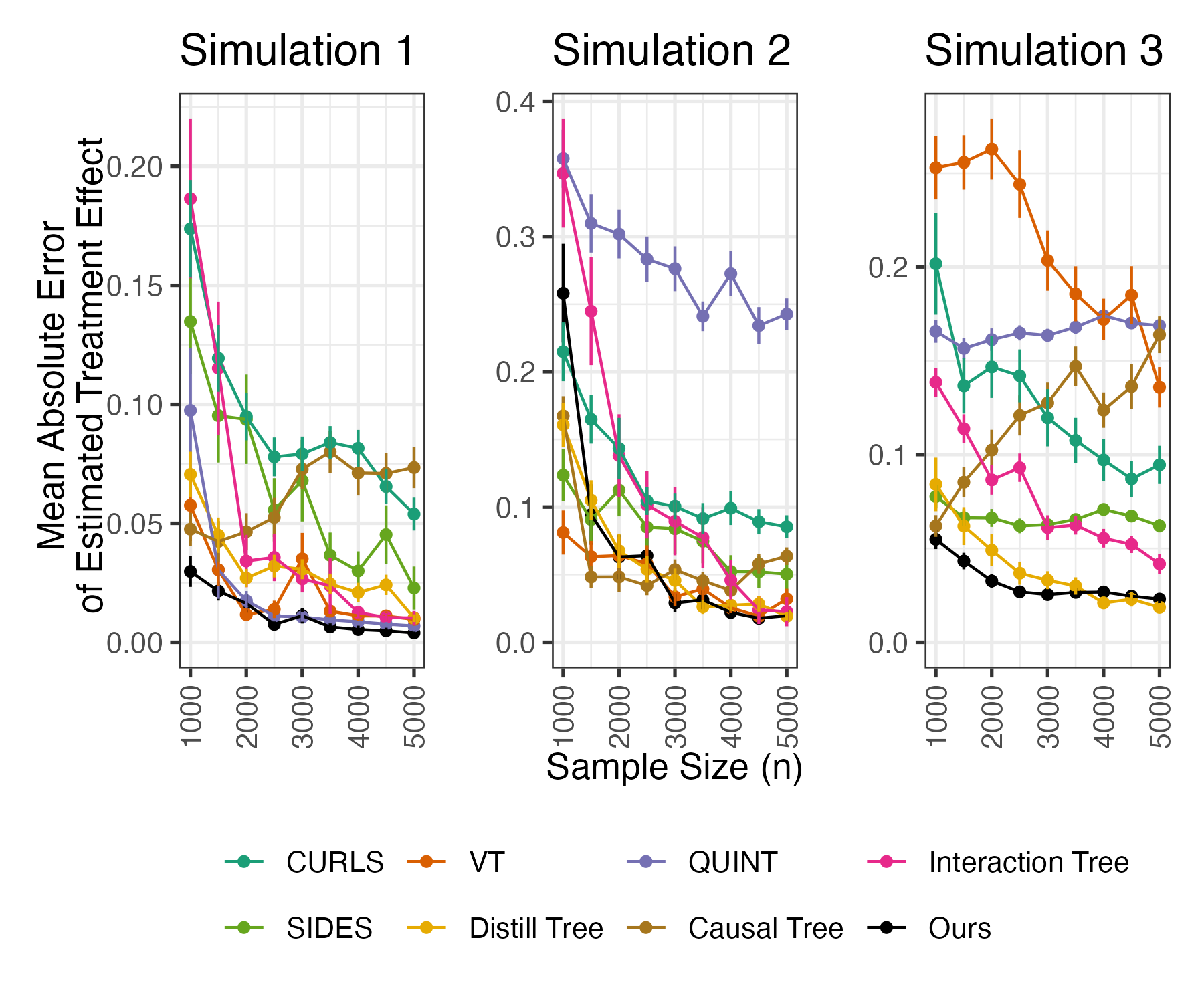} 
    \vspace{-5mm}
    \caption{The sample sizes versus the absolute value of the difference between the estimated treatment effect of the learned subgroup $\hat Q$ and that of the ground-truth subgroup $Q_{gt}$ (lower is better). All simulations are repeated 50 times, and we use the error bar to represent the standard error.}
    \label{fig:synthetic_diff_te}
\end{figure}
\begin{figure}[ht]
    \centering
    \includegraphics[width=0.95\linewidth]{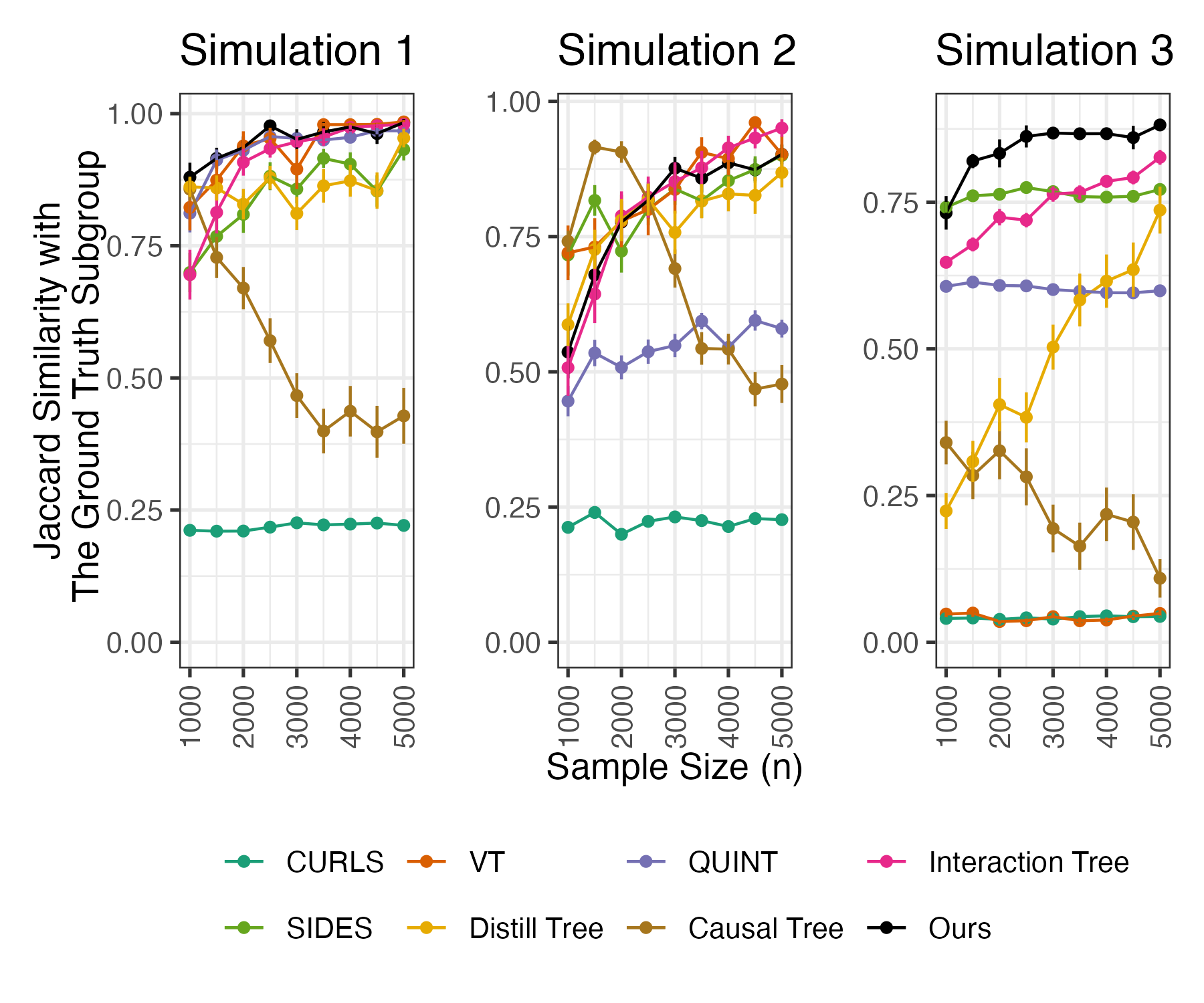}
        \vspace{-5mm}
    \caption{The sample sizes versus the Jaccard similarity of the learned subgroup $\hat Q$ and the ground-truth subgroup $Q_{gt}$ (higher is better). All simulations are repeated 50 times, and we use the error bar to represent the standard error.}
    \label{fig:syntheic_jaccard}
\end{figure}
\vspace{-4mm}
\section{Experiment}
\subsection{Implementation and baseline methods}
% \noindent \textbf{Baseline methods.}
We consider seven baseline methods in total. Five use specifically designed causal heuristics to learn the subgroups, including SIDES~\cite{lipkovich2011subgroup, lipkovich2017tutorial}, Interaction Trees~\cite{su2009subgroup}, QUINT~\cite{dusseldorp2014qualitative}, Causal Tree~\cite{athey2016recursive}, and the recently proposed CURLS~\citep{zhou2024curls}. The remaining two adopt the two-step approach, i.e., first estimating the pointwise conditional treatment effect and then fitting a model to the pointwise estimates to learn the subgroups. These include Virtual Twins (VT)~\citep{foster2011subgroup} (only suitable for binary targets), and the recently proposed Distill Tree~\citep{huang2025distilling}. 

To ensure a fair comparison, we use the same `honest' inference protocol for all methods (i.e., to train the model on a training split, and to estimate the subgroup treatment effects on a held-out test split). 
All hyperparameters are tuned in the same way as in the original work of the baseline method. If the details of hyperparameter tuning are not given, we follow the hyperparameter tuning procedures or use default values in the source code of that original work. We include the full implementation details and hyperparameter tuning in the supplementary materials. 

\subsection{Synthetic Datasets} 
We consider the following simulations: 1) \textbf{(Simulation 1) when features and treatment are independent}, i.e., the features $X_1, X_2 \sim N(0, 1)$, the treatment $T\sim Ber(0.5)$, and the target $Y|X_1 > 1 \wedge T=1 \sim Ber(0.8)$, $Y|X_1 < -1 \wedge T=0 \sim Ber(0.75)$, $Y|otherwise \sim Ber(0.2)$, in which $Ber(.)$ represents the Bernoulli distribution; 2) \textbf{(Simulation 2) when features and treatment are dependent}, i.e., the features $X_1, X_2 \sim N(0, 1)$ and the treatment $T|X_1 >= 0 \sim Ber(0.8)$, $T|X_1 < 0 \sim Ber(0.2)$, and the target $Y$ is generated as in the previous case; 3) \textbf{(Simulation 3) Rule list simulator.} 
The features $X_1,\ldots,X_5 \sim \mathcal{N}(0,1)$ and the treatment $T \sim \mathrm{Ber}(0.5)$. 
Define $\text{rule}_1: X_1 > -1 \,\&\, X_2 > -1 \,\&\, X_3 > -1$, 
$\text{rule}_2: X_1 > -1 \,\&\, X_2 > -1 \,\&\, \neg\text{rule}_1$, 
$\text{rule}_3: X_1 > -1 \,\&\, \neg\text{rule}_1 \,\&\, \neg\text{rule}_2$. 
Then 
$Y|T=1,\text{rule}_1 \sim \mathrm{Ber}(0.8)$, 
$Y|T=1,\text{rule}_2 \sim \mathrm{Ber}(0.6)$, 
$Y|T=1,\text{rule}_3 \sim \mathrm{Ber}(0.4)$, and 
$Y|\text{otherwise} \sim \mathrm{Ber}(0.2)$.

We vary the sample sizes from $1000$ to $5000$, and repeat the simulation $50$ times. We first apply our method to learn the maximum-effect subgroup $\hat{Q}$ from data, and obtain the estimated subgroup treatment effect denoted as $\hat{A}(\hat{Q})$. Next, as we know the ground-truth maximum-effect subgroup, which we denote as $Q_{gt}$, we also investigate its estimated subgroup treatment effect, denoted as $\hat{A}(Q_{gt})$. We report the difference between them $|\hat{A}(\hat Q) - \hat{A}(Q_{gt})|$ (lower is better) in Figure~\ref{fig:synthetic_diff_te}, and meanwhile report the Jaccard similarity between the instances contained in $Q_{gt}$ and those contained in $\hat{Q}$ in Figure~\ref{fig:syntheic_jaccard} (higher is better). 

As shown in Figures \ref{fig:synthetic_diff_te}–\ref{fig:syntheic_jaccard}, our method performers competitive in Simulation 1 and 2 where the ground-truth partition contains only two subgroups. Notably, several heuristics-driven causal methods already begin to fail in these simple cases, either showing suboptimal effect estimates or poor subgroup recovery. In Simulation 3, our method clearly outperforms the five heuristics-driven baselines in both metrics with smaller absolute effect error and higher Jaccard similarity. Further, although the two-step approach algorithm Distill Tree is a close competitor in effect error (Figure~\ref{fig:synthetic_diff_te}), it performs substantially worse in Jaccard similarity (Figure~\ref{fig:syntheic_jaccard}).
% Across all simulations and sample sizes (Figures~\ref{fig:synthetic_diff_te}--\ref{fig:syntheic_jaccard}), our method clearly outperforms the five heuristics-driven baselines in both metrics: it yields smaller absolute effect error and higher Jaccard similarity to the ground-truth subgroup. These results support our general recipe of recasting maximum-effect subgroup discovery as a standard supervised learning task to reveal the data-generating model. 
% Distill Tree is a close competitor in effect error (Figure~\ref{fig:synthetic_diff_te}), but performs substantially worse in Jaccard similarity.
% Our method also consistently outperforms VT, the other baseline with the two-step approach. 

\begin{table}[ht]
  \centering
  \scalebox{0.75}{
  \begin{tabular}{@{} l c c @{}}
    \toprule
    Method
      & \makecell{Mean ground-truth\\subgroup treatment effect}
      & \makecell{Proportions of ranked \\ first among all 77 datasets} \\
    \midrule
  \textbf{Ours} & \textbf{10.540} & \textbf{0.519} \\ 
CURLS & 7.410 & 0.180 \\ 
  CausalTree & 7.843 & 0.143 \\ 
  DistillTree & 7.451 & 0.130 \\ 
  InteractionTree & 6.280 & 0.039 \\ 
  QUINT & 5.135 & 0.000 \\ 
  SIDES & 4.622 & 0.013 \\ 
    \bottomrule
  \end{tabular}
  }
  \caption{Ground-truth subgroup treatment effect for the learned subgroups of the 77 semi-synthetic datasets (higher is better). Our method ranks first in $51.9\%$ of all datasets. }
  \label{table:semi}
\end{table}
\vspace{-5mm}
\subsection{Semi-synthetic Datasets}
We consider the commonly used semi-synthetic simulator ACIC-2016~\citep{dorie2019automated}, which can generate semi-synthetic datasets based on real features from the Infant Health and Development Program (IHDP)~\citep{louizos2017causal}. We used 77 semi-synthetic datasets generated by this simulator, and we again adopt the honest estimation as previously described to learn the maximum-effect subgroup.

Since the ground-truth treatment effect is known for each single instance (as the outcome is given for both $T=1$ and $T=0$), the treatment effect for each individual is known. We take the average treatment effects for all instances in our learned subgroup and report this ground-truth subgroup average treatment effects for each dataset (higher is better). As reported in Table~\ref{table:semi}, our learned subgroups have the highest ground-truth subgroup treatment effect and rank first in $51.9\%$ of all cases. 

\begin{figure}[ht]
    \centering
    \includegraphics[width=0.9\linewidth]{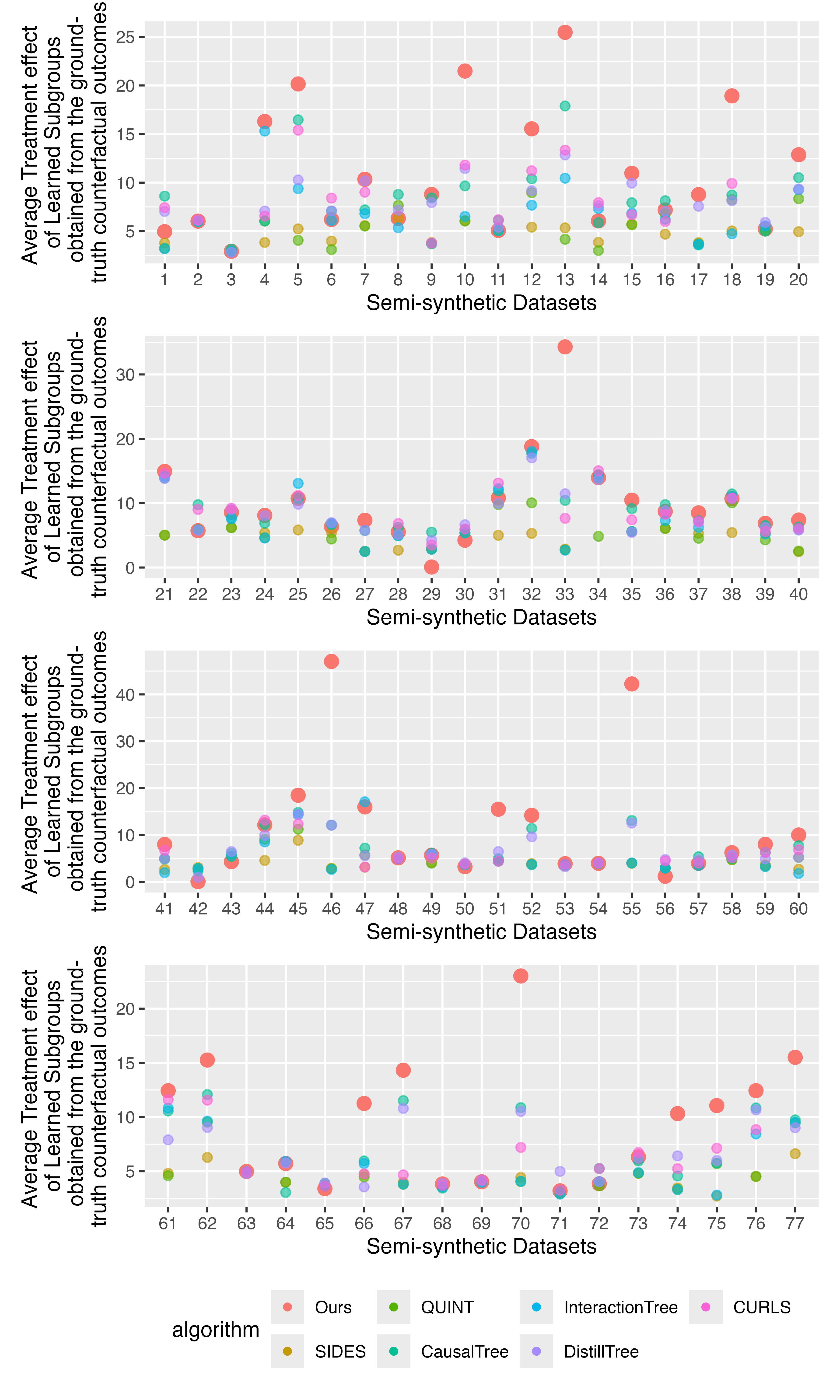}
        \vspace{-3mm}
    \caption{Ground-truth average treatment effect of the learned subgroup (higher is better). }
    \label{fig:semi_full}
\end{figure}

Further, we demonstrate the ground-truth average treatment effect for the learned subgroups for each individual dataset in Figure~\ref{fig:semi_full}. The results show that, for a significant proportion of all 77 datasets, we have identified subgroups with substantially larger average treatment effects than those of other baselines. We finally emphasize that these semi-synthetic datasets are \emph{not} simulated based on a partition-based model; thus, it demonstrates that our method is empirically robust and can generalize to more general cases. 

\section{Conclusion and Discussion}
We studied the problem of learning the subgroup with the maximum treatment effect under the the structural causal model framework. We first showed that any maximizer must exhibit homogeneous pointwise treatment effects, which motivated us to consider the partition-based model. 
Our main theorem then established that discovering the maximum-effect subgroup reduces to a standard regression/classification problem under a partition-based model, and hence challenges the necessity of the bespoke `causal' heuristics. We instantiated the approach with CART and paired it with the `honest' estimation. We compared against several baselines with both synthetic and semi-synthetic datasets and demonstrated that our method has superior performance in discovering maximum-effect subgroups.

Overall, our results support a simple and general recipe: learn a supervised partition, then estimate effects and select the subgroup. The limitations of this work may be the no-hidden-confounder assumption and the exact-partition assumption. Future research directions might include relaxing these assumptions and exploring other (advanced) algorithms for learning partition-based models.

\newpage
\bibliographystyle{plainnat}
\bibliography{aaai2026}

\begin{thebibliography}{43}
\providecommand{\natexlab}[1]{#1}

\bibitem[{Athey and Imbens(2016)}]{athey2016recursive}
Athey, S.; and Imbens, G. 2016.
\newblock Recursive partitioning for heterogeneous causal effects.
\newblock \emph{Proceedings of the National Academy of Sciences}, 113(27):
  7353--7360.

\bibitem[{ATHEY, TIBSHIRANI, and WAGER(2019)}]{athey2019generalized}
ATHEY, S.; TIBSHIRANI, J.; and WAGER, S. 2019.
\newblock GENERALIZED RANDOM FORESTS.
\newblock \emph{The Annals of Statistics}, 47(2): 1148--1178.

\bibitem[{Athey and Wager(2019)}]{athey2019estimating}
Athey, S.; and Wager, S. 2019.
\newblock Estimating treatment effects with causal forests: An application.
\newblock \emph{Observational studies}, 5(2): 37--51.

\bibitem[{Atzmueller(2015)}]{atzmueller2015subgroup}
Atzmueller, M. 2015.
\newblock Subgroup discovery.
\newblock \emph{Wiley Interdisciplinary Reviews: Data Mining and Knowledge
  Discovery}, 5(1): 35--49.

\bibitem[{Breiman et~al.(1984)Breiman, Friedman, Stone, and
  Olshen}]{breiman1984classification}
Breiman, L.; Friedman, J.; Stone, C.~J.; and Olshen, R.~A. 1984.
\newblock \emph{Classification and regression trees}.
\newblock CRC press.

\bibitem[{Brița, van~der Linden, and Demirovi{\'c}(2025)}]{brita2025optimal}
Brița, C.~E.; van~der Linden, J.~G.; and Demirovi{\'c}, E. 2025.
\newblock Optimal Classification Trees for Continuous Feature Data Using
  Dynamic Programming with Branch-and-Bound.
\newblock In \emph{Proceedings of the AAAI Conference on Artificial
  Intelligence}, volume~39, 11131--11139.

\bibitem[{Chen et~al.(2025)Chen, Pan, Gan, and Wang}]{chen2025mosic}
Chen, W.; Pan, W.; Gan, K.; and Wang, F. 2025.
\newblock MOSIC: Model-Agnostic Optimal Subgroup Identification with
  Multi-Constraint for Improved Reliability.
\newblock \emph{arXiv preprint arXiv:2504.20908}.

\bibitem[{Chernozhukov et~al.(2018)Chernozhukov, Chetverikov, Demirer, Duflo,
  Hansen, Newey, and Robins}]{chernozhukov2018double}
Chernozhukov, V.; Chetverikov, D.; Demirer, M.; Duflo, E.; Hansen, C.; Newey,
  W.; and Robins, J. 2018.
\newblock Double/debiased machine learning for treatment and structural
  parameters.

\bibitem[{Clark and Niblett(1989)}]{clark1989cn2}
Clark, P.; and Niblett, T. 1989.
\newblock The CN2 induction algorithm.
\newblock \emph{Machine learning}, 3(4): 261--283.

\bibitem[{Cohen(1995)}]{cohen1995ripper}
Cohen, W.~W. 1995.
\newblock Fast effective rule induction.
\newblock In \emph{Machine learning proceedings 1995}, 115--123. Elsevier.

\bibitem[{Curth and Van~der Schaar(2021)}]{curth2021inductive}
Curth, A.; and Van~der Schaar, M. 2021.
\newblock On inductive biases for heterogeneous treatment effect estimation.
\newblock \emph{Advances in Neural Information Processing Systems}, 34:
  15883--15894.

\bibitem[{Dorie et~al.(2019)Dorie, Hill, Shalit, Scott, and
  Cervone}]{dorie2019automated}
Dorie, V.; Hill, J.; Shalit, U.; Scott, M.; and Cervone, D. 2019.
\newblock Automated versus do-it-yourself methods for causal inference: Lessons
  learned from a data analysis competition.
\newblock \emph{Statistical Science}, 34(1): 43--68.

\bibitem[{Dusseldorp and Van~Mechelen(2014)}]{dusseldorp2014qualitative}
Dusseldorp, E.; and Van~Mechelen, I. 2014.
\newblock Qualitative interaction trees: a tool to identify qualitative
  treatment--subgroup interactions.
\newblock \emph{Statistics in medicine}, 33(2): 219--237.

\bibitem[{Foster, Taylor, and Ruberg(2011)}]{foster2011subgroup}
Foster, J.~C.; Taylor, J.~M.; and Ruberg, S.~J. 2011.
\newblock Subgroup identification from randomized clinical trial data.
\newblock \emph{Statistics in medicine}, 30(24): 2867--2880.

\bibitem[{F{\"u}rnkranz, Gamberger, and
  Lavra{\v{c}}(2012)}]{furnkranz2012foundations}
F{\"u}rnkranz, J.; Gamberger, D.; and Lavra{\v{c}}, N. 2012.
\newblock \emph{Foundations of rule learning}.
\newblock Springer Science \& Business Media.

\bibitem[{Goligher et~al.(2023)Goligher, Lawler, Jensen, Talisa, Berry,
  Lorenzi, McVerry, Chang, Leifer, Bradbury et~al.}]{goligher2023heterogeneous}
Goligher, E.~C.; Lawler, P.~R.; Jensen, T.~P.; Talisa, V.; Berry, L.~R.;
  Lorenzi, E.; McVerry, B.~J.; Chang, C.-C.~H.; Leifer, E.; Bradbury, C.;
  et~al. 2023.
\newblock Heterogeneous treatment effects of therapeutic-dose heparin in
  patients hospitalized for COVID-19.
\newblock \emph{Jama}, 329(13): 1066--1077.

\bibitem[{Hahn, Murray, and Carvalho(2020)}]{hahn2020bayesian}
Hahn, P.~R.; Murray, J.~S.; and Carvalho, C.~M. 2020.
\newblock Bayesian regression tree models for causal inference: Regularization,
  confounding, and heterogeneous effects (with discussion).
\newblock \emph{Bayesian Analysis}, 15(3): 965--1056.

\bibitem[{Hu, Rudin, and Seltzer(2019)}]{hu2019optimal}
Hu, X.; Rudin, C.; and Seltzer, M. 2019.
\newblock Optimal sparse decision trees.
\newblock \emph{Advances in Neural Information Processing Systems}, 32.

\bibitem[{Huang, Tang, and Kenney(2025)}]{huang2025distilling}
Huang, M.; Tang, T.~M.; and Kenney, A.~M. 2025.
\newblock Distilling heterogeneous treatment effects: Stable subgroup
  estimation in causal inference.
\newblock \emph{arXiv preprint arXiv:2502.07275}.

\bibitem[{K{\"u}nzel et~al.(2019)K{\"u}nzel, Sekhon, Bickel, and
  Yu}]{kunzel2019metalearners}
K{\"u}nzel, S.~R.; Sekhon, J.~S.; Bickel, P.~J.; and Yu, B. 2019.
\newblock Metalearners for estimating heterogeneous treatment effects using
  machine learning.
\newblock \emph{Proceedings of the national academy of sciences}, 116(10):
  4156--4165.

\bibitem[{Lavra{\v{c}} et~al.(2004)Lavra{\v{c}}, Kav{\v{s}}ek, Flach, and
  Todorovski}]{lavravc2004subgroup}
Lavra{\v{c}}, N.; Kav{\v{s}}ek, B.; Flach, P.; and Todorovski, L. 2004.
\newblock Subgroup discovery with CN2-SD.
\newblock \emph{Journal of Machine Learning Research}, 5(Feb): 153--188.

\bibitem[{Lee et~al.(2025)Lee, Liu, Song, Li, and Zhang}]{lee2025subgroupte}
Lee, S.; Liu, R.; Song, W.; Li, L.; and Zhang, P. 2025.
\newblock SubgroupTE: Advancing Treatment Effect Estimation with Subgroup
  Identification.
\newblock \emph{ACM transactions on intelligent systems and technology}, 16(3):
  1--23.

\bibitem[{Lipkovich, Dmitrienko, and
  B~D'Agostino~Sr(2017)}]{lipkovich2017tutorial}
Lipkovich, I.; Dmitrienko, A.; and B~D'Agostino~Sr, R. 2017.
\newblock Tutorial in biostatistics: data-driven subgroup identification and
  analysis in clinical trials.
\newblock \emph{Statistics in medicine}, 36(1): 136--196.

\bibitem[{Lipkovich et~al.(2011)Lipkovich, Dmitrienko, Denne, and
  Enas}]{lipkovich2011subgroup}
Lipkovich, I.; Dmitrienko, A.; Denne, J.; and Enas, G. 2011.
\newblock Subgroup identification based on differential effect search—a
  recursive partitioning method for establishing response to treatment in
  patient subpopulations.
\newblock \emph{Statistics in medicine}, 30(21): 2601--2621.

\bibitem[{Loh, Cao, and Zhou(2019)}]{loh2019subgroup}
Loh, W.-Y.; Cao, L.; and Zhou, P. 2019.
\newblock Subgroup identification for precision medicine: A comparative review
  of 13 methods.
\newblock \emph{Wiley Interdisciplinary Reviews: Data Mining and Knowledge
  Discovery}, 9(5): e1326.

\bibitem[{Louizos et~al.(2017)Louizos, Shalit, Mooij, Sontag, Zemel, and
  Welling}]{louizos2017causal}
Louizos, C.; Shalit, U.; Mooij, J.~M.; Sontag, D.; Zemel, R.; and Welling, M.
  2017.
\newblock Causal effect inference with deep latent-variable models.
\newblock \emph{Advances in neural information processing systems}, 30.

\bibitem[{Nagpal et~al.(2020)Nagpal, Wei, Vinzamuri, Shekhar, Berger, Das, and
  Varshney}]{nagpal2020interpretable}
Nagpal, C.; Wei, D.; Vinzamuri, B.; Shekhar, M.; Berger, S.~E.; Das, S.; and
  Varshney, K.~R. 2020.
\newblock Interpretable subgroup discovery in treatment effect estimation with
  application to opioid prescribing guidelines.
\newblock In \emph{Proceedings of the ACM Conference on Health, Inference, and
  Learning}, 19--29.

\bibitem[{Nie and Wager(2021)}]{nie2021quasi}
Nie, X.; and Wager, S. 2021.
\newblock Quasi-oracle estimation of heterogeneous treatment effects.
\newblock \emph{Biometrika}, 108(2): 299--319.

\bibitem[{Pearl(2009)}]{pearl2009causality}
Pearl, J. 2009.
\newblock \emph{Causality}.
\newblock Cambridge university press.

\bibitem[{Pearl(2012)}]{pearl2012calculus}
Pearl, J. 2012.
\newblock The do-calculus revisited.
\newblock \emph{arXiv preprint arXiv:1210.4852}.

\bibitem[{Rosenbaum and Rubin(1983)}]{RosenbaumRubin1983}
Rosenbaum, P.~R.; and Rubin, D.~B. 1983.
\newblock {The central role of the propensity score in observational studies
  for causal effects}.
\newblock \emph{Biometrika}, 70(1): 41--55.

\bibitem[{Rothwell(2005)}]{rothwell2005subgroup}
Rothwell, P.~M. 2005.
\newblock Subgroup analysis in randomised controlled trials: importance,
  indications, and interpretation.
\newblock \emph{The Lancet}, 365(9454): 176--186.

\bibitem[{Rubin(1974)}]{Rubin1974}
Rubin, D.~B. 1974.
\newblock {Estimating causal effects of treatments in randomized and
  nonrandomized studies}.
\newblock \emph{Journal of Educational Psychology}, 66(5): 688--701.

\bibitem[{Seibold, Zeileis, and Hothorn(2016)}]{seibold2016model}
Seibold, H.; Zeileis, A.; and Hothorn, T. 2016.
\newblock Model-based recursive partitioning for subgroup analyses.
\newblock \emph{The international journal of biostatistics}, 12(1): 45--63.

\bibitem[{Shi, Blei, and Veitch(2019)}]{shi2019adapting}
Shi, C.; Blei, D.; and Veitch, V. 2019.
\newblock Adapting neural networks for the estimation of treatment effects.
\newblock \emph{Advances in neural information processing systems}, 32.

\bibitem[{Su et~al.(2009)Su, Tsai, Wang, Li et~al.}]{su2009subgroup}
Su, X.; Tsai, C.-L.; Wang, H.; Li, B.; et~al. 2009.
\newblock Subgroup analysis via recursive partitioning.
\newblock \emph{Journal of Machine Learning Research}, 10(2).

\bibitem[{Therneau et~al.(2015)Therneau, Atkinson, Ripley, and
  Ripley}]{therneau2015package}
Therneau, T.; Atkinson, B.; Ripley, B.; and Ripley, M.~B. 2015.
\newblock Package ‘rpart’.
\newblock \emph{Available online: cran. ma. ic. ac.
  uk/web/packages/rpart/rpart. pdf (accessed on 20 April 2016)}, 2: 5--32.

\bibitem[{Van~Leeuwen and Knobbe(2012)}]{van2012diverse}
Van~Leeuwen, M.; and Knobbe, A. 2012.
\newblock Diverse subgroup set discovery.
\newblock \emph{Data Mining and Knowledge Discovery}, 25(2): 208--242.

\bibitem[{Wang and Rudin(2022)}]{wang2022causal}
Wang, T.; and Rudin, C. 2022.
\newblock Causal rule sets for identifying subgroups with enhanced treatment
  effects.
\newblock \emph{INFORMS journal on computing}, 34(3): 1626--1643.

\bibitem[{Yang and van Leeuwen(2022)}]{yang2022truly}
Yang, L.; and van Leeuwen, M. 2022.
\newblock Truly unordered probabilistic rule sets for multi-class
  classification.
\newblock In \emph{Joint European Conference on Machine Learning and Knowledge
  Discovery in Databases}, 87--103. Springer.

\bibitem[{Yang and van Leeuwen(2024)}]{yang2024conditional}
Yang, L.; and van Leeuwen, M. 2024.
\newblock Conditional density estimation with histogram trees.
\newblock \emph{Advances in Neural Information Processing Systems}, 37:
  117315--117339.

\bibitem[{Zhang et~al.(2017)Zhang, Le, Liu, Zhou, and Li}]{zhang2017mining}
Zhang, W.; Le, T.~D.; Liu, L.; Zhou, Z.-H.; and Li, J. 2017.
\newblock Mining heterogeneous causal effects for personalized cancer
  treatment.
\newblock \emph{Bioinformatics}, 33(15): 2372--2378.

\bibitem[{Zhou et~al.(2024)Zhou, Yang, Liu, Gu, Sun, and Chen}]{zhou2024curls}
Zhou, J.; Yang, L.; Liu, X.; Gu, X.; Sun, L.; and Chen, W. 2024.
\newblock CURLS: Causal Rule Learning for Subgroups with Significant Treatment
  Effect.
\newblock In \emph{Proceedings of the 30th ACM SIGKDD Conference on Knowledge
  Discovery and Data Mining}, 4619--4630.

\end{thebibliography}

\makeatletter
\@ifundefined{isChecklistMainFile}{
  % We are compiling a standalone document
  \newif\ifreproStandalone
  \reproStandalonetrue
}{
  % We are being \input into the main paper
  \newif\ifreproStandalone
  \reproStandalonefalse
}
\makeatother

\ifreproStandalone
\documentclass[letterpaper]{article}
\usepackage[submission]{aaai2026}
\setlength{\pdfpagewidth}{8.5in}
\setlength{\pdfpageheight}{11in}
\usepackage{times}
\usepackage{helvet}
\usepackage{courier}
\usepackage{xcolor}
\frenchspacing

\begin{document}
\fi
\setlength{\leftmargini}{20pt}
\makeatletter\def\@listi{\leftmargin\leftmargini \topsep .5em \parsep .5em \itemsep .5em}
\def\@listii{\leftmargin\leftmarginii \labelwidth\leftmarginii \advance\labelwidth-\labelsep \topsep .4em \parsep .4em \itemsep .4em}
\def\@listiii{\leftmargin\leftmarginiii \labelwidth\leftmarginiii \advance\labelwidth-\labelsep \topsep .4em \parsep .4em \itemsep .4em}\makeatother

\setcounter{secnumdepth}{0}
\renewcommand\thesubsection{\arabic{subsection}}
\renewcommand\labelenumi{\thesubsection.\arabic{enumi}}

\newcounter{checksubsection}
\newcounter{checkitem}[checksubsection]

\newcommand{\checksubsection}[1]{%
  \refstepcounter{checksubsection}%
  \paragraph{\arabic{checksubsection}. #1}%
  \setcounter{checkitem}{0}%
}

\newcommand{\checkitem}{%
  \refstepcounter{checkitem}%
  \item[\arabic{checksubsection}.\arabic{checkitem}.]%
}
\newcommand{\question}[2]{\normalcolor\checkitem #1 #2 \color{blue}}
\newcommand{\ifyespoints}[1]{\makebox[0pt][l]{\hspace{-15pt}\normalcolor #1}}

\section*{Reproducibility Checklist}

% \vspace{1em}
% \hrule
% \vspace{1em}

% \textbf{Instructions for Authors:}

% This document outlines key aspects for assessing reproducibility. Please provide your input by editing this \texttt{.tex} file directly.

% For each question (that applies), replace the ``Type your response here'' text with your answer.

% \vspace{1em}
% \noindent
% \textbf{Example:} If a question appears as
% %
% \begin{center}
% \noindent
% \begin{minipage}{.9\linewidth}
% \ttfamily\raggedright
% \string\question \{Proofs of all novel claims are included\} \{(yes/partial/no)\} \\
% Type your response here
% \end{minipage}
% \end{center}
% you would change it to:
% \begin{center}
% \noindent
% \begin{minipage}{.9\linewidth}
% \ttfamily\raggedright
% \string\question \{Proofs of all novel claims are included\} \{(yes/partial/no)\} \\
% yes
% \end{minipage}
% \end{center}
% %
% Please make sure to:
% \begin{itemize}\setlength{\itemsep}{.1em}
% \item Replace ONLY the ``Type your response here'' text and nothing else.
% \item Use one of the options listed for that question (e.g., \textbf{yes}, \textbf{no}, \textbf{partial}, or \textbf{NA}).
% \item \textbf{Not} modify any other part of the \texttt{\string\question} command or any other lines in this document.\\
% \end{itemize}

% You can \texttt{\string\input} this .tex file right before \texttt{\string\end\{document\}} of your main file or compile it as a stand-alone document. Check the instructions on your conference's website to see if you will be asked to provide this checklist with your paper or separately.

% \vspace{1em}
% \hrule
% \vspace{1em}

% The questions start here

\checksubsection{General Paper Structure}
\begin{itemize}

\question{Includes a conceptual outline and/or pseudocode description of AI methods introduced}{(yes/partial/no/NA)}
Yes

\question{Clearly delineates statements that are opinions, hypothesis, and speculation from objective facts and results}{(yes/no)}
Yes

\question{Provides well-marked pedagogical references for less-familiar readers to gain background necessary to replicate the paper}{(yes/no)}
Yes

\end{itemize}
\checksubsection{Theoretical Contributions}
\begin{itemize}

\question{Does this paper make theoretical contributions?}{(yes/no)}
Yes

	\ifyespoints{\vspace{1.2em}If yes, please address the following points:}
        \begin{itemize}
	
	\question{All assumptions and restrictions are stated clearly and formally}{(yes/partial/no)}
	Yes

	\question{All novel claims are stated formally (e.g., in theorem statements)}{(yes/partial/no)}
	Yes

	\question{Proofs of all novel claims are included}{(yes/partial/no)}
	Yes

	\question{Proof sketches or intuitions are given for complex and/or novel results}{(yes/partial/no)}
	Yes

	\question{Appropriate citations to theoretical tools used are given}{(yes/partial/no)}
	Yes

	\question{All theoretical claims are demonstrated empirically to hold}{(yes/partial/no/NA)}
	Yes

	\question{All experimental code used to eliminate or disprove claims is included}{(yes/no/NA)}
	Yes
	
	\end{itemize}
\end{itemize}

\checksubsection{Dataset Usage}
\begin{itemize}

\question{Does this paper rely on one or more datasets?}{(yes/no)}
Yes

\ifyespoints{If yes, please address the following points:}
\begin{itemize}

	\question{A motivation is given for why the experiments are conducted on the selected datasets}{(yes/partial/no/NA)}
	Yes

	\question{All novel datasets introduced in this paper are included in a data appendix}{(yes/partial/no/NA)}
	NA

	\question{All novel datasets introduced in this paper will be made publicly available upon publication of the paper with a license that allows free usage for research purposes}{(yes/partial/no/NA)}
	NA

	\question{All datasets drawn from the existing literature (potentially including authors' own previously published work) are accompanied by appropriate citations}{(yes/no/NA)}
	Yes

	\question{All datasets drawn from the existing literature (potentially including authors' own previously published work) are publicly available}{(yes/partial/no/NA)}
	Yes

	\question{All datasets that are not publicly available are described in detail, with explanation why publicly available alternatives are not scientifically satisficing}{(yes/partial/no/NA)}
	NA

\end{itemize}
\end{itemize}

\checksubsection{Computational Experiments}
\begin{itemize}

\question{Does this paper include computational experiments?}{(yes/no)}
Yes

\ifyespoints{If yes, please address the following points:}
\begin{itemize}

	\question{This paper states the number and range of values tried per (hyper-) parameter during development of the paper, along with the criterion used for selecting the final parameter setting}{(yes/partial/no/NA)}
	Yes

	\question{Any code required for pre-processing data is included in the appendix}{(yes/partial/no)}
	Yes

	\question{All source code required for conducting and analyzing the experiments is included in a code appendix}{(yes/partial/no)}
	Yes

	\question{All source code required for conducting and analyzing the experiments will be made publicly available upon publication of the paper with a license that allows free usage for research purposes}{(yes/partial/no)}
	Yes
        
	\question{All source code implementing new methods have comments detailing the implementation, with references to the paper where each step comes from}{(yes/partial/no)}
	Yes

	\question{If an algorithm depends on randomness, then the method used for setting seeds is described in a way sufficient to allow replication of results}{(yes/partial/no/NA)}
	Yes

	\question{This paper specifies the computing infrastructure used for running experiments (hardware and software), including GPU/CPU models; amount of memory; operating system; names and versions of relevant software libraries and frameworks}{(yes/partial/no)}
	Yes

	\question{This paper formally describes evaluation metrics used and explains the motivation for choosing these metrics}{(yes/partial/no)}
	Yes

	\question{This paper states the number of algorithm runs used to compute each reported result}{(yes/no)}
	Yes

	\question{Analysis of experiments goes beyond single-dimensional summaries of performance (e.g., average; median) to include measures of variation, confidence, or other distributional information}{(yes/no)}
	Yes

	\question{The significance of any improvement or decrease in performance is judged using appropriate statistical tests (e.g., Wilcoxon signed-rank)}{(yes/partial/no)}
	No

	\question{This paper lists all final (hyper-)parameters used for each model/algorithm in the paper’s experiments}{(yes/partial/no/NA)}
	Yes

\end{itemize}
\end{itemize}
\ifreproStandalone
\end{document}
\fi
\newpage

\section{Supplementary Materials}
Due to page limit, we include some additional proofs, implementation details, and additional experiment results in the supplementary materials. 

\section{Appendix I: Additional Proofs}
We first present the proof for Proposition~\ref{prop1} which is skipped in the main body due to the space limit. We then extend Theorem~2 in the main body to the case where hidden confounders do exit.  
% \begin{proposition} \label{prop1}
% $A(Q) = E_{X|X \in Q} P(Y=1|T=1, X) - P(Y=1|T=0, X)$. 
% \end{proposition}

\smallskip \noindent \textbf{Proposition 1.} $A(Q) = E_{X|X \in Q} P(Y=1|T=1, X) - P(Y=1|T=0, X)$. 
\begin{proof}
We first show that $A(Q) = E_{X|X \in Q} A(X)$.
{\small
    \begin{align}
        &P(Y=1 |do(T:=1), X \in Q) - \notag \\
        & P(Y=1 |do(T:=0), X \in Q) \notag \\
        = & \sum_{x \in Q}P(Y=1|do(T:=1), X=x) P(X=x|X\in Q) - \notag \\ 
         &  \sum_{x \in Q}P(Y=1|do(T:=0), X=x) P(X=x|X\in Q) \label{eq:use_insertion_rule}\\
        = & E_{X|X \in Q} A(X) \notag
    \end{align}
}
Next, under the assumption of now hidden confounder, we have $P(Y=1|do(T=t), X=x) = P(Y=1|T=t, X=x), t \in \{0, 1\}$, which means we can drop the `do' in Eq.~\ref{eq:use_insertion_rule} above. Thus, 
{\small
    \begin{align*}
        & E_{X|X \in Q} A(X) \\
        = & \sum_{x \in Q}P(Y=1|do(T:=1), X=x) P(X=x|X\in Q) -  \\ 
         &  \sum_{x \in Q}P(Y=1|do(T:=0), X=x) P(X=x|X\in Q) \\
        = & \sum_{x \in Q}P(Y=1|T=1, X=x) P(X=x|X\in Q) -  \\ 
         &  \sum_{x \in Q}P(Y=1|T=0, X=x) P(X=x|X\in Q) \\
        = & E_{X|X \in Q} P(Y=1|T=1, X) - P(Y=1|T=0, X),
    \end{align*}
}
which completes the proof (note that if $X$ is continuous we can just  change the summation to integral in the proof). 
\end{proof}

\begin{figure}[ht]
\centering
\begin{tikzpicture}[->,>=stealth, node distance=2cm, thick]
    \node (X) at (0,0) {X};
    \node (T) at (2,0) {T};
    \node (Y) at (4,0) {Y};
    \node (S) at (1,-1.5) {S};
    \node (U) at (2,1.5) {U};

    \draw (X) -- (T);
    \draw (T) -- (Y);
    \draw (X) -- (S);
    \draw (S) -- (Y);
    \draw (U) -- (T);
    \draw (U) -- (Y);

\end{tikzpicture}
\caption{DAG for the case where the hidden confounder exists. }\label{fig:dag2}
\end{figure}
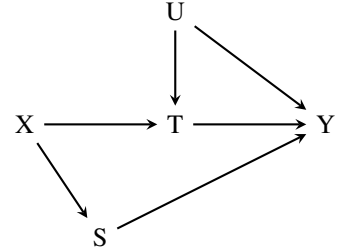
We next consider extending Theorem~2 in the main body of the paper to the case where the hidden confounder exists. We again assume that $Y$ is binary (but can be extended to continuous by replacing $P(.)$ with $E(.)$), $X$ and $U$ discrete (but can be extended to continuous by replacing $\sum$ with the integral sign). 

\setcounter{theorem}{2}%
\begin{theorem}
Consider the corresponding SCM:
\begin{align*}
    X = N_X \notag \\
    U = N_U \notag \\
    T = f_T(X, U, N_T) \notag \\
    Y = f_Y(T, \sum_i^I i \cdot \mathbf{1}_{K_i}(X), U, N_Y) \\
    K_i \subseteq \mathcal{X}, \forall i, j \in I, K_i \cap K_j = \emptyset, \\
    N_X, N_T, N_Y \text{  are independent exogenous noises,} 
\end{align*} 
in which $U$ is a hidden confounder. Denote $S = \sum_{i\in I} i \cdot \mathbf{1}_{K_i}(X), K_i \subseteq \mathcal{X}, \forall i, j \in I, K_i \cap K_j = \emptyset$. Hence our SCM can be represented as the DAG in Figure~\ref{fig:dag2}.

Without loss of generality, assume that
{\small
    \begin{align} \label{eq:assumption2}
       & P(Y=1|S=1, do(T:=1)) - P(Y=1|S=1, do(T:=0)) \geq \notag \\ 
       & P(Y=1|S = i, do(T:=1)) - P(Y=1|S=i, do(T:=0)) \notag \\ 
       & (\forall i \in \{2, ..., n\}) . 
    \end{align} 
}
Then, $\forall Q \subseteq \mathcal{X}$, we have
{\small
    \begin{align} \label{eq:thm2}
        P(Y=1|S=1, do(T:=1)) - 
        P(Y=1|S=1, do(T:=0)) \geq \notag \\ P(Y=1|X\in Q, do(T:=1)) -  P(Y=1|X\in Q, do(T:=0))
    \end{align}  
}
\end{theorem}

\begin{proof}

Without loss of generality, we assume that $S \neq 0$ no matter what value $X$ takes. This can be achieved by making the last subgroup $K_n = \mathcal{X} \setminus (\cup_{i \in \{1, ..., n-1\}} K_i)$. 
The right hand side (RHS) of Eq.~\ref{eq:thm2} is 
\begin{align*}
 \text{RHS} & = \sum_u P(Y=1, U=u|X\in Q, do(T:=1)) - \\
 & P(Y=1, U=u|X\in Q, do(T:=0)) \\
 & = \sum_u \Bigl(\frac{ P(Y=1, U=u, X \in Q| do(T:=1))}{ P(X \in Q, U=u| do(T:=1))} - \\
    & \frac{ P(Y=1, U=u, X \in Q| do(T:=0))}{ P(X \in Q, U=u| do(T:=0))}\Bigr) \\
      & = \sum_u \Bigl( \frac{ \sum_{i} P(Y=1, U=u, X \in Q\cap K_i| do(T:=1))}{ P(X \in Q| do(T:=1))} - \\
      & \frac{  \sum_{i} P(Y=1, U=u, X \in Q\cap K_i| do(T:=0))}{ P(X \in Q| do(T:=0))} \Bigr) \\
      &  = \sum_u \sum_{i} \Bigl(\frac{P(X \in Q \cap K_i | do(T:=1))}{P(X\in Q |do(T:=1))} \cdot \\
      & P(Y=1, U=u|X\in Q\cap K_i, do(T:=1)) \Bigr) - \\
\end{align*}
\begin{align*}
      & \sum_u \sum_{i}  \Bigl( \frac{P(X \in Q \cap K_i | do(T:=0))}{P(X\in Q |do(T:=0))} \cdot \\
      & P(Y=1, U=u|X\in Q\cap K_i, do(T:=0))  \Bigr) \\ 
      & \stackrel{\text{(a)}}{=} \sum_u \sum_i \Bigl(P(Y=1, U=u|X \in Q \cap K_i, do(T:=1)) - \notag\\
      & P(Y=1, U=u|X \in Q \cap K_i, do(T:=0)) \Bigr) \frac{P(X \in Q \cap K_i)}{P(X \in Q)} \notag \\
      = & \sum_i \Bigl(P(Y=1|X \in Q \cap K_i, do(T:=1)) - \notag\\
      & P(Y=1, |X \in Q \cap K_i, do(T:=0)) \Bigr) \frac{P(X \in Q \cap K_i)}{P(X \in Q)} \notag \\
       & \stackrel{}{=}    \sum_i \frac{P(X \in Q \cap K_i)}{P(X \in Q)} \bigl( P(Y=1|S=i, do(T:=1)) -\notag \\
       & P(Y=1|S=i, do(T:=0))\bigr)  \notag\\
       & \stackrel{}{\leq} \sum_i \frac{P(X \in Q \cap K_i)}{P(X \in Q)} \bigl( P(Y=1|S=1, do(T:=1)) - \notag\\ 
       & P(Y=1|S=1, do(T:=0))\bigr)  \notag \\
       & = \text{LHS of Eq.~\ref{eq:thm2}}, \notag 
\end{align*}
where in \\
(a) Note that  $P(X\in Q, U=u| do(T:=0)) = P(X \in Q, U=u)$, since $Y$ blocks the paths between $X \& T$ and $U \& T$ after removing $X \rightarrow T$ and $U \rightarrow T$ (and similarly $P(X\in Q| do(T:=0)) = P(X \in Q)$). 
\end{proof}

\section{Appendix II: Implementation Details}
We next describe the implementation details for our method and all baselines to provide the information for full reproducibility. 

\subsection{Semi-synthetic datasets}
We use the simulator from the Github repository \url{https://github.com/vdorie/aciccomp} to simulate the data for Atlantic Causal Inference Conference Competition. The simulation is based on a real-world dataset, and can use 77 different parameter settings to simulate 77 datasets. We then one-hot encode all categorical features. The final datasets all have 57 feature variables and 1 continuous target variable. 

\subsection{Our method and baselines}
\begin{itemize}
    \item Ours: we adopt the `rpart' package~\citep{therneau2015package} to implement CART and leverage the built-in cross-validation for pruning the tree. There is no need to specify the hyperparameter range as the built-in function will iteratively prune the weakest node and return the corresponding coefficients before the model complexity term. We only need to use those coefficients for cross-validation for pruning the tree. We then implement ourselves the `honest' inference for estimating the treatment effect. 
    \item CURLS: we leverage the author's implementation of the code, which can be found at \url{https://osf.io/zwp2k/}. We follow the hyper-parameter tuning guideline specified in the original paper~\citep{zhou2024curls}: i.e., to use cross-validation to tune rule length $L \in \{3, 4, 5, 6\}$, and variance weight $\lambda \in \{0.1, ... , 1.5\}$. 
    \item Distill Tree~\citep{huang2025distilling}: we use the author's implementation, which can be found at: \url{https://tiffanymtang.github.io/causalDT/index.html}. As suggested by the default options, we picked the teacher model as ``causal forest" and student model as ``decision tree" (they also use `rpart'). We prune the `rpart' decision tree by specifying the prune option as `min', which is essentially the same way of pruning as in our own method. 
    \item Interaction Tree: We adopt the code shared by the author of the original paper~\citep{su2009subgroup}, which can be found at \url{https://drive.google.com/file/d/1XUIWapeSE3m5VdteTxmyYgSP3A3eeNy4/view}. This is tree-based method. The tree was first grown and then pruned with a held-out validation set, as suggested by the original author in a example of illustrating how to use the source code. 
    \item QUINT: we use the R package (also named as ``quint") developed by the authors of the original paper~\citep{dusseldorp2014qualitative}, which can be found at \url{https://cran.r-project.org/web/packages/quint/}. We set the bootstrap option, which is used for correcting bias, as `True' as suggested by the R package documents. Other hyper-parameters are not so critical and we left as defaults (e.g., the minimum size of each left). 
    \item Causal Tree: we use the R package named ``causalTree" which can be found at \url{https://github.com/susanathey/causalTree}. As suggested by the default values, the `honest' split and `honest' validation are both used. The proposed criterion `CT'~\citep{athey2016recursive} is used both for splitting and for cross-validation. 
    \item Virtual Twin: We use the implementation from the R package named 'aVirutalTwins', which can be found at \url{https://cran.r-project.org/web/packages/aVirtualTwins/index.html}. Hyper-parameters and model options are set as suggested by the `README' file of the package (e.g., the number of trees for the random forest is set as 500). 
    \item SIDES: We use the implementation from the R package `RSIDES', which can be found at \url{https://cran.r-project.org/package=rsides}. As mentioned in the `Related Work' section, SIDES provides several options for the users to decide on the splitting criterion, which can be specified via the `analysis method' parameters. We pick the `Z-test for proportions' for binary target $Y$ and 'T-test' for continuous $Y$. Following the default settings on other parameters, we set the search width as $2$, search depth as the number of columns (larger than the default value), min subgroup size as 10, and number of permutations for multiple correction adjustment as $10$. 
\end{itemize}

\section{Appendix III: Additional Experiment Results}
We report the significance testing results on our experiment results. 
\begin{table}[ht]
\centering
\caption{Significance test results on synthetic datasets, comparing subgroup discovery methods using two evaluation metrics: Jaccard similarity between learned and ground-truth subgroups, and the Difference between the estimated and true treatment effects within the identified subgroup (Diff. TE). All p-values are computed using the one-sided Wilcoxon rank-sum test, and Holm-adjusted p-values are reported for multiple comparison correction.}
\scalebox{0.85}{
\begin{tabular}{lcccc}
\toprule
\multirow{2}{*}{Method} & 
\multicolumn{2}{c}{Jaccard similarity} & 
\multicolumn{2}{c}{Diff. TE} \\
\cmidrule(lr){2-3} \cmidrule(lr){4-5}
 & p-value & p\_holm & p-value & p\_holm \\
\midrule
CURLS            & 7.451e-09 & 5.215e-08 & 2.235e-08 & 1.118e-07 \\
SIDES            & 3.339e-04 & 1.002e-03 & 1.104e-05 & 4.417e-05 \\
VT               & 1.401e-02 & 1.401e-02 & 2.678e-04 & 5.357e-04 \\
Distill Tree     & 2.287e-06 & 9.149e-06 & 6.263e-04 & 6.263e-04 \\
QUINT            & 7.451e-08 & 4.470e-07 & 1.490e-08 & 8.941e-08 \\
Causal Tree      & 1.542e-06 & 7.711e-06 & 1.182e-04 & 3.545e-04 \\
Interaction Tree & 1.354e-03 & 2.708e-03 & 7.451e-09 & 5.215e-08 \\
\bottomrule
\end{tabular}
}
\end{table}

\begin{table}[ht]
\centering
\caption{Significance test results on semi-synthetic datasets, comparing methods using the ground-truth treatment effect of the learned maximum-effect subgroup. All p-values are computed using the one-sided Wilcoxon rank-sum test, and Holm-adjusted p-values are reported for multiple comparison  correction.}
\begin{tabular}{lcc}
\toprule
Method & p-value & p\_holm \\
\midrule
CURLS            & 6.505e-06  & 1.301e-05 \\
Causal Tree      & 1.208e-04  & 1.208e-04 \\
Distill Tree     & 4.235e-06  & 1.271e-05 \\
Interaction Tree & 2.170e-12  & 8.679e-12 \\
QUINT            & 1.292e-14  & 7.750e-14 \\
SIDES            & 1.528e-14  & 7.750e-14 \\
\bottomrule
\end{tabular}
\end{table}

\end{document}